\documentclass[sigconf]{acmart}
\usepackage[T1]{fontenc}
\usepackage{graphicx}
\usepackage{amsmath}
\usepackage{textcomp}
\usepackage{booktabs}
\usepackage{float}

\usepackage{hyperref}

\hypersetup{
    colorlinks=true,          % Enable colored links
    linkcolor=black,          % Color of internal links
    citecolor=black,          % Color of citations
    urlcolor=black,           % Color of external URLs
    pdftitle={Do Language Models create world models?},    % Title of the document
    pdfauthor={Klimkowski, Suaiter, Ortiz de Zarate, Feuerstein},   % Author of the document
    pdfkeywords={Large Language Models, World Models, Text Embeddings.}, % Keywords for the document
    pdfproducer={LaTeX},     % Producer of the document
    pdfcreator={pdflatex}    % Creator of the document
}
\usepackage{color}

\usepackage[all]{hypcap} % 13/5/2025 - Para que me lleve a la figura si hago click en la referencia (caso contrario me lleva a la caption) - Fede

\acmConference[]{}{}{}
\acmBooktitle{}

\title[Recovering Temporal and Geographic Signals from LLM Embeddings]{Recovering Temporal and Geographic Signals from Language Model Embeddings}
\author{
Esteban Feuerstein \orcid{0000-0003-2985-810X}  
 \and Victoria Klimkowski 
 \and  Juan Manuel Ortiz de Zarate \orcid{0000-0002-0291-1997}
 \and Federico Hernán Suaiter \orcid{0009-0005-3932-9422} 
 }

 \affiliation{
 \institution{Departamento de Computación, Facultad de Ciencias Exactas y Naturales, Universidad de Buenos Aires }
 \and 
\institution{Instituto de Ciencias de la Computación, CONICET-UBA}
\city{Buenos Aires} \country{Argentina}}
\AtBeginDocument{%
  \fancyhead[RE]{Recovering Temporal and Geographic Signals from LLM Embeddings}
  \fancyhead[LO]{E. Feuerstein, V. Klimkowski, JM Ortiz de Zárate, and FH Suaiter}
}

\begin{document}



\begin{abstract}
Understanding whether language-model embeddings encode structured real-world information is important for both representation analysis and information retrieval. We study this question for temporal and geographic signals using a simple projection-based method that operates directly on output embeddings. Given a small set of seed examples, the method defines an axis in embedding space and ranks texts or entities by their projection onto that axis.

Our approach is fully black-box and model-agnostic: it requires only embeddings, without access to model weights, internal activations, auxiliary probes, or additional training. This makes it applicable to modern embedding models available only through APIs and provides a lightweight way to analyze whether temporal and spatial dimensions are present in their representation spaces.

We apply the method to temporal and geographic datasets and find that embedding projections recover meaningful chronological and spatial structure. These results provide evidence that output embeddings encode signals relevant to time and space, while also offering a practical tool for interpretability and for downstream temporal and geographic information retrieval tasks, such as temporal ordering, geographic ranking, and tagging.
\end{abstract}

% \begin{abstract}
% We find new evidence that LLMs generate grounded representations of the real world, through the application of a simple and general technique that consists in projecting the embedded representation of a given text along different dimensions. 
% In particular, we apply this method to analyze the temporal and geographic dimensions. 

% The only input our technique requires as input are the embeddings generated by the language model, making it an extremely simple, black-box approach. 

% A distinctive quality of our technique is that it only requires as input the embeddings generated by the language model. 
% No probes on the internal activations nor any access to the neural networks are required, as well as no or additional models nor training are needed. 
% This allows us to evaluate in a fast, simple, and explicit manner
% the intrinsic capabilities of language representation models to capture these dimensions, which is another important value of our technique.

% Besides this theoretical application to establish LLM's capabilities, the method can also be used to improve interpretability by analyzing how the models represent or capture other concepts, and to improve core temporal and geographic information retrieval tasks, such as temporal and geographic tagging.
% \end{abstract}

\maketitle              % typeset the header of the contribution

% \renewcommand{\shortauthors}{E. Feuerstein, V. Klimkowski, JM Ortiz de Zárate, and FH Suaiter}
% \makeatletter
% \renewcommand{\@shorttitle}{Recovering Temporal and Geographic Signals from LLM Embeddings}
\makeatother
\section{Introduction}
\label{introduction}
% Una introducción que hable de los LLMs en general y de su enorme auge, y de todo lo que se está descubriendo

Large Language Models (LLMs) have introduced a new perspective from which to analyze the topics of intelligence, knowledge, and the handling of language by computers. In recent years, many variants and improvements have been developed, from early models~\cite{radford2018improving,devlin2018bert} to current models with hundreds of billions of parameters trained over petabytes of data. For an extensive review of the current landscape of LLMs, the reader can refer to~\cite{eisenstein2019introduction,qiu2020pre,han2021pre,zhao2023survey,minaee2024large}.

% Párrafo que habla de habilidades emergentes + tiempo y espacio en el paper de Gurnee y Tegmark
Recently, there has been much interest in a series of works focused on what are referred to as \textit{emergent abilities}, which are defined --paraphrasing~\cite{wei2022emergent}-- as ``\textit{abilities that are not present in smaller models but are present in larger models}''. In turn, following~\cite{anderson1972more},``emergence'' is ``\textit{when quantitative changes in a system result in qualitative changes in behavior}''. Among these abilities, what stands out is the capacity of LLMs, not only of \textit{learning} a large number of patterns or statistical properties from the texts, but to \textit{generate} grounded representations that reflect the real world. In~\cite{gurnee2023language}, the authors find evidence that this is the case by analyzing the learned representations of spatial and temporal datasets in the Llama-2 family of models, discovering that LLMs learn linear representations of space and time across multiple scales and that these representations are robust to prompt variations and unified across different entity types. Notably, they identify individual \textit{space neurons} and \textit{time neurons} that reliably encode spatial and temporal coordinates.
More recently,~\cite{DBLP:conf/iclr/EngelsMLGT25} extended the vision to find that some of these learned representations are inherently multidimensional, and provide an updated multi-dimensional superposition hypothesis to account for these kind of features. It also develops ``a theoretically grounded
and empirically practical test that uses sparse autoencoders to find irreducible features."

% Explicacion de como es la idea general 
This work takes a further step in the verification of linear feature representations (namely, time and space coordinates), based on 
%a series of findings and experiments obtained by applying 
a simple methodology 
that takes as input the embeddings generated by the language representation model, and hence does not require access to internal layers of the neural network. 
We first verify the basic hypothesis that language models create temporal models. Specifically, we establish a \textit{temporal dimension} for a corpus, and then map the texts of the corpus onto that dimension, finding that the projection yields a rather accurate ordering. 
We then deepen the investigation by addressing granularity and testing different ways in which the temporal signal may be established.
The spatial hypothesis is verified in a similar way, although with further challenges due to the intrinsic three-dimensionality of the ground world they represent. In summary, we provide further evidence that current language representation models generate embeddings that contain sound temporal and geographic information. 

%by generating embeddings of texts corresponding to extremes of a given temporal interval.
%We deepen this investigation by addressing granularity as well, testing robustness of the temporal signal in a flow of daily news, where semantic noise can often overpower temporal drift. We established the \textit{temporal dimension} as the geometrical difference between these two extremes. We then map the other texts of the corpus to that dimension, finding that the projection results in a rather accurately ordered sequence. To validate that the signal emerges from the embedding structure and not from chance, we contrasted our extremes method against both a random control and ordered random seeds which demonstrates the consistency of the results.

% EXPLICAR COMO SE DIFERENCIA DE TRABAJOS PREVIOS
Prior work has shown that language models encode temporal and spatial information, either by analyzing internal activations and training probes~\cite{gurnee2023language}, or by using task-specific temporal supervision, as in TempoBERT~\cite{rosin2022time}. Our specific contribution is a lightweight, black-box method that requires only the embeddings produced by the language representation model. Thus, we do not require access to model weights, internal activations, auxiliary probes, or additional training. This enables a fast, simple, and transparent evaluation of temporal and spatial signals in modern embedding models, including those available only through APIs. Furthermore, by casting temporal text classification as a ranking/ordering problem~\cite{niculae2014temporal}, our approach naturally connects to information retrieval settings and avoids committing to a fixed set of temporal classes.

In addition to serving as another indication of the emergent capability of word embeddings to establish what has been referred to as~\textit{``a set of more coherent and grounded representations that reflect the real world''}~\cite{gurnee2023language}, 
this work contributes to LLM interpretability research by enabling the analysis of whether LLMs correctly model abstract concepts, such as time and space. It also provides another insight into how LLMs learn those concepts, but in an easier, more natural, and more explicit way than probing-based analyses, which require access to internal activations or weights~\cite{gurnee2023language}. Our contribution is not the invention of projection-based or \textit{d}-ness scoring itself~\cite{bolukbasi2016man,waller2021quantifying}, nor the first claim that temporal and spatial signals can emerge in language models. Rather, our contribution is to apply this idea to text embeddings produced by language representation models: while prior uses of \textit{d}-ness scoring focused on different embedding objects, such as community embeddings derived from user co-occurrence graphs~\cite{waller2021quantifying}, we show that the same projection-based principle can recover temporal and geographic signals directly from language-model output embeddings. 
Using a simple procedure that operates exclusively on model outputs, we recover meaningful temporal orderings and geospatial regularities without training auxiliary probes, without relying on task-specific temporal supervision such as TempoBERT~\cite{rosin2022time}, and without inspecting network internals. This makes the approach broadly applicable in black-box settings---e.g., when models are only accessible through an API---and complements prior white-box and probe-based analyses by showing that time and space signals can be detected directly from output embeddings.

Last, but not least, our findings can potentially be used to improve some of the downstream tasks that are part of 
Temporal and Geographic Information Retrieval (T-IR and G-IR),
%the Temporal Information Retrieval (T-IR) and Geographic Information Retrieval (G-IR) areas, 
as temporal and spatial information can be inferred from the data. For more on T-IR and G-IR, see~\cite{campos2014survey} and the book~\cite{purves2018geographic}, respectively. %In~\cite{zhu2023large}, the authors provide an extensive review of the intersection between LLMs and IR systems, covering key aspects such as query rewriting, retrieval, reranking, and reading components.
An extensive review of the intersection between LLMs and IR systems—covering key aspects such as query rewriting, retrieval, reranking, and reading—is provided in~\cite{zhu2023large}.

The rest of this paper is organized as follows: in~\autoref{sec:related-work} we summarize related research. In Sections~\ref{sec:method} and~\ref{sec:experimental-results} we present our method and experimental results, while~\autoref{sec:conclusions} is devoted to discussion, conclusions and further research directions.
The code of this method has been published along with the customized datasets used in this work\footnote{https://github.com/FedericoSu/BIICC-2024}.
% https://github.com/FedericoSu/BIICC-2024
\section{Related Work}
\label{sec:related-work}
% EMERGENT ABILITIES

\textit{Word representation} is a fundamental area of research in natural language processing. The progress of machine learning techniques in recent years enabled the development of newer and more complex models that are capable of learning higher-quality vector representations from massive datasets containing billions of words~\cite{mikolov2013efficient,pennington2014glove,bojanowski2017enriching,radford2018improving,devlin2018bert,reimers2019sentence}. In~\cite{reimers2019sentence}, SBERT is presented, a modification of BERT~\cite{devlin2018bert} that obtains high-quality sentence embeddings that can be compared using cosine similarity at a low computational cost without losing BERT's precision. SBERT adds a pooling operation to BERT's output to obtain a vector representation of the entire sentence, and fine-tuning is performed 
%on a \textit{Siamese architecture}, 
so that the produced embeddings are semantically meaningful and can be compared using cosine similarity. 
%For this, the objective function minimizes the squared error between the \textit{gold similarity} and the cosine similarity (inferred).
Another extension of BERT is  TempoBERT~\cite{rosin2022time}, a temporally adapted language model trained with a time-masking objective to improve temporal awareness in contextual representations. Unlike our setting, it relies on task-specific temporal supervision and model training to encode temporal information. Our approach instead asks whether temporal signals can be recovered directly from off-the-shelf output embeddings, without additional training or access to model internals.

Word vector representations capture syntactic and semantic regularities in a very simple way~\cite{mikolov2013efficient,bojanowski2017enriching}. By using basic arithmetic operations and cosine distance measures over word vector representations, it is possible to capture various syntactic and semantic relationships, despite the lack of explicit supervision. These regularities not only solve analogy tasks~\cite{mikolov2013linguistic,levy2014linguistic,vylomova2015take,gladkova2016analogy} or position adjectives on a dimensional scale~\cite{kim2013deriving}, but also enable studies on biases and analyses of cultural and social dimensions in the field of sociology~\cite{bolukbasi2016man,garg2018word,kozlowski2019geometry,waller2021quantifying}. These works are based on the hypothesis that difference vectors have an intrinsic meaning, as their direction and magnitude encode semantic and syntactic relationships. By carefully selecting the initial vectors, it is possible to generate vector representations that capture various cultural and social dimensions, such as gender, wealth, age, and partisanship. This positions words and documents along the chosen dimension $d$, obtaining a \textit{d-ness} score for each document based on the projection of each vector with respect to that dimension vector, resulting in an inherent ranking.

Previous research~\cite{lou09language,lou12representing} has shown that language encodes geographic information. In~\cite{cohn2024evaluating} it was found that, while none of the tested models were able to reliably reason about cardinal directions, all models exhibited some spatial reasoning abilities. Additionally,~\cite{lietard2021language} shows that language models encode only limited geographic information, with larger models generally performing better. As noted there, modern language models encode some syntactic~\cite{tenney2019bert} and semantic knowledge~\cite{reif2019negotiation}, as well as real-world knowledge. For example, models have been shown to capture commonsense knowledge~\cite{dson2019commonsense}, function as knowledge bases~\cite{petroni2019language}, and store a range of factual information~\cite{jiang2020can,roberts2020much}.
In the setting of 
Linguistic Spatial Models, prior work such as,~\cite{DBLP:conf/ijcnlp/KonkolBNH17}
has shown that natural language encodes geographic information, and that relative coordinates can
be approximately recovered with techniques like multidimensional scaling, co-occurrence statistics, or probing. The aim there is to evaluate the quality of particular word embeddings. 

A pioneer work on modeling temporal information
for the automatic dating of documents is~\cite{ec22721d8ee1475cb9a2306d72ab3939}, where unigram language models were used  to
classify Dutch texts through normalised log-likelihood
ratio (NLLR). Other works also proposed using lexical features for automatic dating 
\cite{dalli-wilks-2006-automatic, 5693378, 10.1145/2063576.2063892}

As discussed in the introduction,~\cite{gurnee2023language} shows that LLMs form world (spatial and temporal) models. The primary distinction from that work lies in our use of \textit{$d$-ness scoring} --which allows us to analyze the embeddings outputted by the models instead of their internal weights-- and \textit{augmentation} methods, originally proposed in~\cite{bolukbasi2016man}. These methods have been revisited in~\cite{waller2021quantifying}, reinterpreting these scores as actual dimensions rather than biases, as suggested in~\cite{kozlowski2019geometry}. To the best of our knowledge, this is the first time \textit{d-ness} score has been applied to the temporal and geographic dimensions. Its application to text embeddings is also original to our work, as~\cite{waller2021quantifying} used it for a completely different kind of embeddings: community embeddings in a user co-occurrence graph. \textit{d-ness}, text, and communities where considered in~\cite{demarco2023measuring}, which introduced a text-based technique to quantify the alignment of online communities along social dimensions by generating text embeddings that are used to score communities on different axes of the political-ideological spectrum.
\section{Method}
\label{sec:method}

% CONTRIBUCION + ESTRUCTURA DE LA SECCION

    Our methodological contribution lies in introducing an approach to measuring the ability of language models to capture temporal and geospatial information from textual data. We hypothesize that events, time expressions, topics, locations, monuments, and cities provide sufficient information for the language model to position the resulting embeddings in a vector space that exhibits regularities in both temporal and geospatial dimensions, resembling the semantic, syntactic, and social regularities observed in previous works~\cite{mikolov2013linguistic,kim2013deriving,levy2014linguistic,vylomova2015take,gladkova2016analogy,kozlowski2019geometry}. We begin by outlining the general scoring algorithm designed to generate the temporal and geographical dimensions, presented first in \cite{waller2021quantifying} based on the work in \cite{bolukbasi2016man,kozlowski2019geometry}. After that, we detail the specific sampling and partitioning strategies used to validate the different signals. Next, we describe the evaluation metrics used to assess the capacity of various language models to represent temporal and geographic dimensions. Finally, we summarize the  language models considered during this research.

\subsection{d-ness Scoring}
% CÓMO SE CALCULAN LOS PUNTAJES

To identify a dimension $d_0$ using embedding models, the basic method consists of selecting a seed pair of vectors $(p, q)$ that (ideally) differ only in the dimension under study. Then we calculate the projection of the vector representation of a document $c$ on dimension $d_0$ using cosine similarity ($ S_{C}$). In other words, we define:
\begin{equation}
    d_0 = p - q 
 \qquad \qquad
    d_0\text{-ness}(c) = S_{C}(c, d_0)
\end{equation}

This method can be extended for a more robust definition of the dimension --and we do so in this work-- by choosing multiple seed pairs in a set $P=\{(p_1, p_2)\}$ with $K=|P|$ and defining:

\begin{equation}
d = \sum_{(p_1, p_2) \in P}{p_1 - p_2}
\end{equation}

To interpret this method, we can rewrite the formula using the distributive property of the inner product with respect to addition, and considering normalized vectors $c$ ($||c|| = 1$) we get:
% \begin{align*}
% d\text{-ness}(c) &= \frac{c \cdot \sum_{(p_1, p_2) \in P}{(p_1 - p_2)}}{||d||} \\
%                  &= \frac{(\sum_{(p_1, p_2) \in P}{c \cdot p_1}) - (\sum_
%                  {(p_1, p_2) \in P}{c \cdot p_2})}{||d||} \\
% \end{align*}
\begin{equation}
d\text{-ness}(c) = \frac{(\sum_{(p_1, p_2) \in P}{c \cdot p_1}) - (\sum_{(p_1, p_2) \in P}{c \cdot p_2})}{||d||}
\end{equation}

Note that this is equivalent to averaging over the first and second components of the $K$ pairs and then computing the difference between those two values.
We observe that, ultimately, a high value of $d$-ness for a vector $c$ indicates that $c$ is similar to the first components of the pairs in $P$, but different from the second components of those pairs. However, recall that we aim to choose the pairs in $P$ in such a way that the components of each pair are similar in all dimensions but  $d$; therefore, the only way for $c$ to be more similar to $p_1$ than to $p_2$ is precisely due to its value in the underlying dimension that differentiates $p_1$ from $p_2$. For this reason, we would typically choose the $K$ seed pairs that comprise P as extreme opposites on the $d$ dimension. As we will see in the following, this basic method can be further extended to capture in different ways the signal we are trying to capture.

\subsection{Seed Selection Strategies}
\label{samplingstrategies}

To distinguish whether the temporal signal emerges from a genuine representation of time or merely from random high-dimensional noise, we introduce three distinct strategies for selecting the seed sets $P$. These strategies are applied specifically to validate the robustness of the signal in high-density datasets.

\textit{Extremes} method: We select the first $K$ elements of the time period as the \textit{early} set ($P_1$) and the last $K$ elements as the \textit{late} set ($P_2$). This method aims to capture the global direction of semantic change.

\textit{Random Control (Total Random)}: We select $2K$ elements completely at random from the dataset. These are arbitrarily assigned to the \textit{early} or \textit{late} sets without regard for their actual chronological timestamp. This serves as a baseline; if the temporal signal is real, this method should yield near-zero correlations.

\textit{Random Sorted (Ordered Random)}: We select $2K$ random elements from the timeline, but we respect their relative chronological order. The randomly selected elements are sorted by date; the first $K$ are assigned to the \textit{early} set and the remaining $K$ to the \textit{late} set. This validates whether the arrow of time is distributed throughout the entire dataset or if it is an artifact exclusive to the absolute extremes of the corpus.

\subsection{Partitioning (Chunking):}

To analyze how the length of the temporal interval affects the temporal signal, we employ a \textit{chunking} strategy. We partition the total time period of our dataset into $m$ disjoint, contiguous segments (chunks).
By varying $m$, we manipulate the length of the time window. For $m=1$, the axis is constructed using the entire dataset. As $m$ increases, the time window decreases. Within each chunk, we apply the sampling strategies described above, using $K$ seeds. This allows us to measure whether the temporal signal stabilizes as the days-per-chunk increase.

\subsection{Evaluation Metrics}
% CÓMO SE EVALÚAN LOS EXPERIMENTOS

The scores produced by the $d$-ness scoring method naturally result in a ranking, allowing us to use for comparison ranking similarity measures such as Kendall's $\tau$~\cite{kendall1938new} and  Pearson's correlation coefficient ($r$). This approach is particularly well-suited here because it emphasizes the relative order of items, making it independent of the absolute values of the scores. By comparing rankings instead of raw scores, we can reliably evaluate results across different models and datasets without concerns about scale differences.

Kendall's $\tau$ correlation  quantifies the compatibility between two  rankings. Its values range from -1 (strong disagreement) to 1 (strong agreement). A value of 0 indicates an uncorrelated or  \textit{random} relationship. Let $C$ be the number of concordant pairs (i.e., pairs for which both rankings share the same relative order), and let $D$ represent the number of discordant pairs. With that, we define
$    \tau = 2\frac{C - D}{n(n-1)}.$

\subsection{Tested Models}

Voyage AI, OpenAI, and Google Gemini\footnote{\url{https://docs.voyageai.com/docs/introduction},~ {\url{https://platform.openai.com/docs/overview}},~\url{https://ai.google.dev/gemini-api/docs} respectively}
provide representation models that generate text embeddings, in the spirit of SBERT~\cite{reimers2019sentence}, which can be accessed in a black-box manner through an API.

Across our experiments, we consider a range of models and only report full results in the figures and tables where relevant. As representative endpoints of this range, we use (i) OpenAI \texttt{ada-002} (8191-token context window, 1536-dimensional embeddings) and (ii) Google Gemini \texttt{text-embedding-004} (2048-token context window, 768-dimensional embeddings).

For high-density analysis, we also incorporate additional models, including voyage-large-2, voyage-3-large, multilingual-2, embed-english-v3.0, embed-english-v2.0, embed-multilingual-v3.0 and embed-multilingual-v2.0.

\section{Experimental Results}
\label{sec:experimental-results}
%- Confidence intervals (e.g., 95%) may be included when comparing models to determine whether the observed differences are statistically meaningful.
%- The authors might consider Spearman’s correlation in addition to Kendall’s correlation as an alternative metric for assessing ranking-based relationships in their analysis.
%- The discussion on partisan differences in model behavior (Republican vs. Democratic speeches) is interesting but underexplored. A more detailed analysis of the differences between these two sets of speeches would improve the interpretation of the results.

In this section, we detail the experimental setup and the results obtained through the method described in~\autoref{sec:method} across various language representation models and datasets. The findings demonstrate the models' effectiveness in capturing both the temporal and the geographic dimensions encoded in the datasets.

\subsection{Time Dimension}

For our analysis, we prepared a dataset to quantify the ability of language models to capture the temporal dimension, focusing on both the document's creation time and focus time (the period or moment the document refers to). Another focal point was to take into account high-frequency data, such as news articles, to analyze the impact in the results when there is semantic noise involved. In the following paragraphs, we provide a more detailed presentation of these datasets and the preprocessing steps we carried out to ensure the reliability and consistency of our analyses, as well as other data sources used. We remark that additional datasets were thoroughly tested with similar results, but are not included in this article for space reasons.
    
\subsubsection{Datasets}

\paragraph{Presidential Speeches.} 
This dataset is a subset of the presidential speeches delivered in the U.S. Specifically, it includes the \textit{State of the Union Address} 
%(\textit{SOTU}) 
speeches from the period 1988-2019. %These are annual events in the U.S. where the sitting president reports to Congress on the state of the country. 
The dataset is freely accessible and available through the \textit{The American Presidency Project} website\footnote{\label{ref:american-project}\url{https://www.presidency.ucsb.edu/about}}.
%This dataset has notable properties. 
The speeches in this dataset have a clearly defined creation time and focus, both aligned with the year of delivery. Their length is sufficient to generate embeddings that capture temporal information, which is essential as attention-based models perform best with at least 800 tokens for time prediction tasks \cite{ray2019ad3}. 
%Although the creation time and focus align in this dataset, further research is needed to understand the implications when they do not. 
    
\paragraph{Party Affiliations.} To ensure that the seed pair only differs in the temporal dimension, we also experiment with grouping speeches by party. To do so, we used an additional dataset that contains each president's party affiliation, along with the start and end dates of their terms\footnote{Also  publicly available at \url{https://www.presidency.ucsb.edu/about}}. Using the party affiliation of the president for the period corresponding to the speech delivery date, we generated two subsets of data: \textbf{Conservative Speeches}, corresponding to speeches delivered by presidents affiliated with the Republican Party, and \textbf{Democratic Speeches}, corresponding to speeches delivered by presidents affiliated with the Democratic Party. We will refer to the dataset that includes all presidential speeches under analysis as \textbf{Presidential} \textbf{Speeches}.

\paragraph{High-density News.} We constructed a dataset using news articles from the \textit{World} section of a major newspaper, spanning from May 2020 to October 2025. This data has daily granularity, with multiple articles per day. We generated a \textit{daily embedding}  by averaging the vectors of all articles (concatenating title and subtitle) from each date.

\subsubsection{Methodology Analysis}
For the first time-related experiments, with the Presidential Speeches datasets, we used the basic seed selection strategies, only varying the $K$ parameter, that is, the number of seed pairs that are averaged to get the $d$ dimension.
In the case of high density news, to distinguish genuine temporal signals from random noise, we contrasted the three seed selection methods described in \autoref{samplingstrategies}.

%\textit{Extremes Method:} Selecting the first $K$ and last $K$ elements from a time period to define the axis. 
%\textit{Random Sorted:} Selecting $2K$ random elements throughout a time period but respecting their chronological order when assigning them to the \textit{early} or \textit{late} sets
%\textit{Random Control (Total Random):} by selection $K$ random elements for early and $K$ random elements for late without taking into account their relative order. 

\subsubsection{Results}
\begin{figure}[htb]
\begin{minipage}{.48\textwidth}
    \centering
    \includegraphics[width=\textwidth]{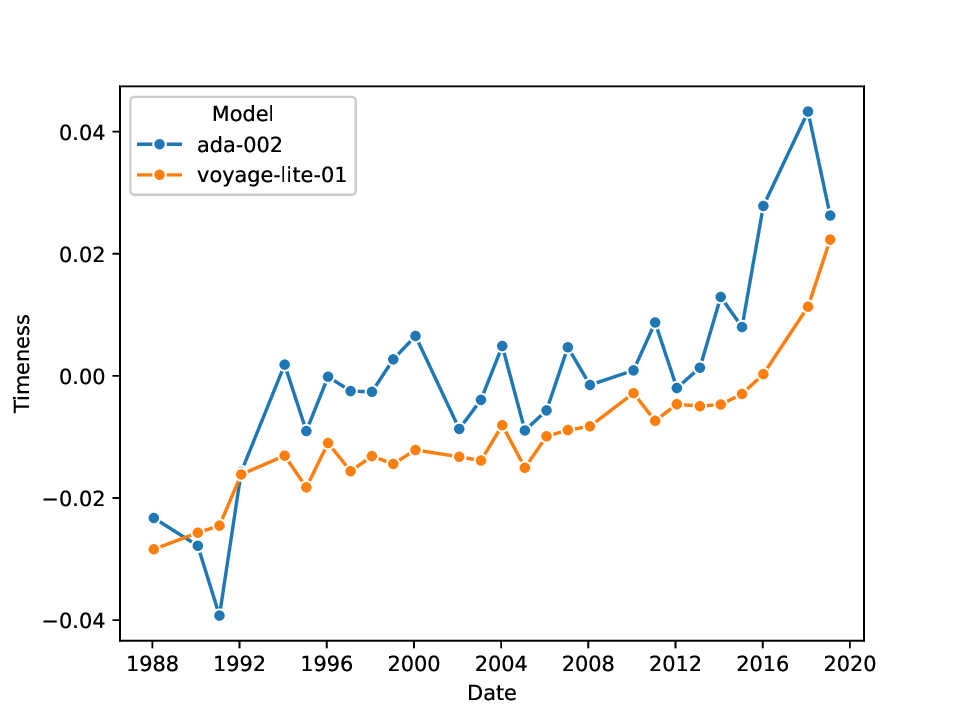}
    \caption{\textbf{Varying models}. \textit{Timeness scores vs. actual time of delivery of the speeches for two models, K = 3.}}
    \label{fig:time-timeness-models}
\end{minipage}\hfill
\begin{minipage}{.48\textwidth}
    \centering
    \includegraphics[width=\textwidth]{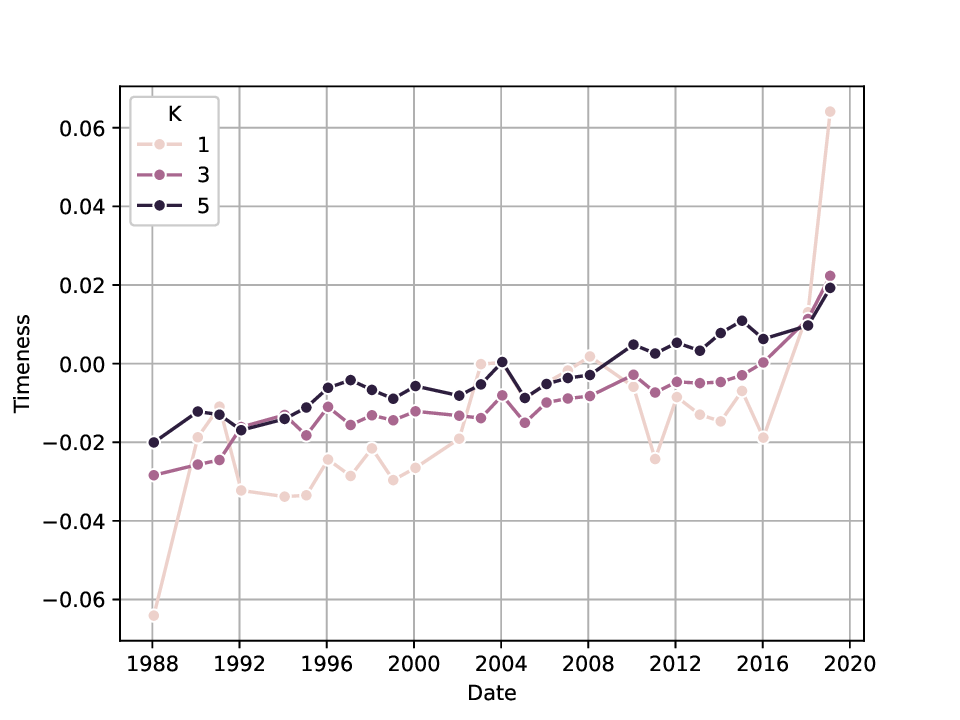}
    \Description{}
    \caption{\textbf{Varying K}. \textit{Timeness scores vs. actual time of delivery of the speeches for different values of the K hyperparameter for the \texttt{voyage-lite-01} model.}}
    \label{fig:time-timeness-k}
\end{minipage}
 \end{figure}

We initially applied our method to compute the d-ness score  corresponding to the time dimension (\textit{timeness}) to the Presidential Speeches dataset. We selected $K$ seed pairs by identifying the $K$ most widely time-separated speech pairs, ensuring no speech is used more than once. To embed the speeches into a vector, we simply truncate them up to the maximum token window allowed by the representation model. In \autoref{fig:time-timeness-models}, we observe that both \texttt{voyage-lite-01} and \texttt{ada-002} capture temporal information, and the scores obtained highly correlate with the actual time of delivery of the speeches. Besides, we compare the performance of both models  (using $K = 3$). 
In \autoref{fig:time-timeness-k}, we can observe how  varying the hyperparameter $K$  impacts the scores obtained with the \texttt{voyage-lite-01} model. Specifically, we see that for $K = 1$, the scores are highly unstable over time, while for higher values, we obtain more stable results. This suggests that we need to take $K\geq 3$ for our scores to be robust enough to be relied on as a time metric. 

\begin{figure}[!htb]
    \centering
    \includegraphics[width=0.525\textwidth]{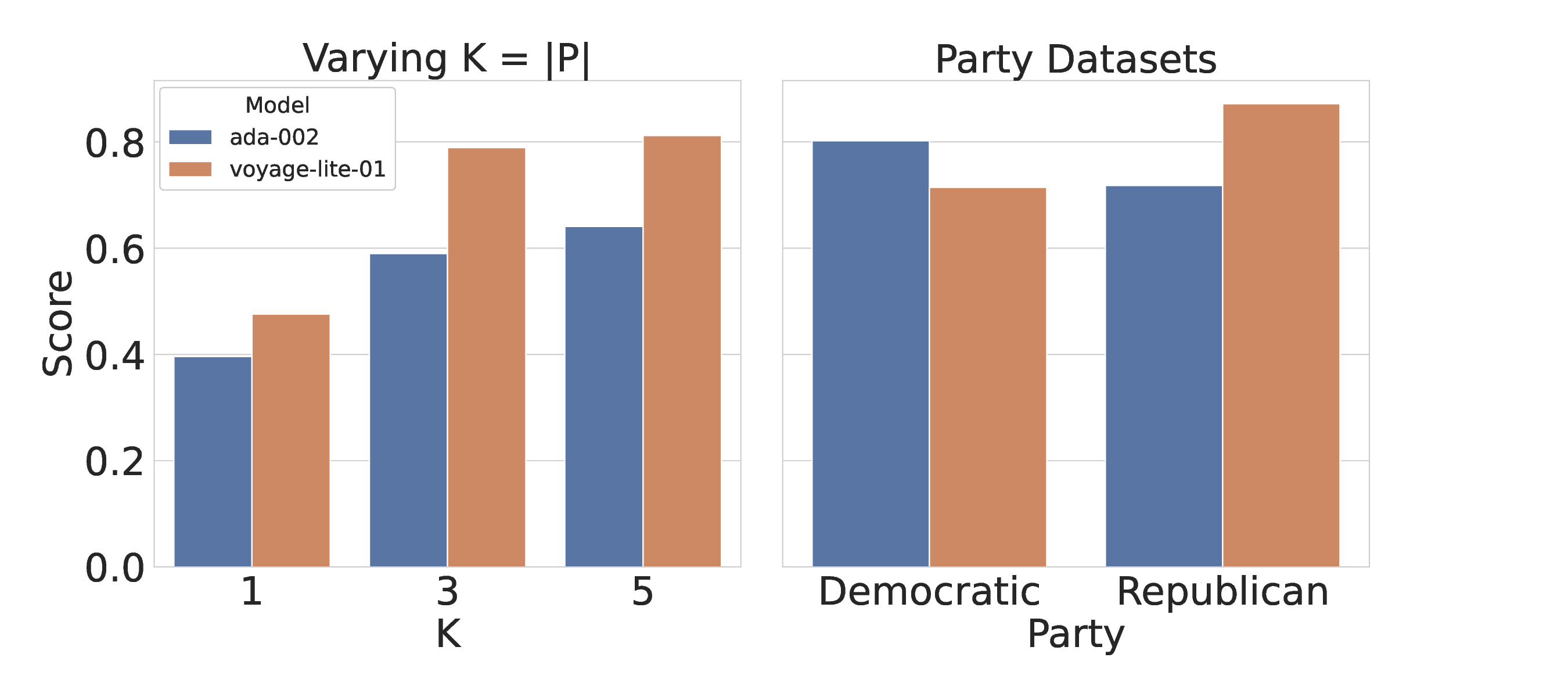}
    \Description{}
    \caption{Temporal modeling evaluation. Left: correlation vs. $K$. Right: Kendall’s $\tau$ by party (with $K=3$).}
   % \caption{Evaluation of the ability of language models to model the temporal dimension. The left chart shows the scores varying with the K parameters. The right chart shows the correlation obtained by applying the method, segregating the speeches by party affiliation (using K = 3), and employing Kendall's $\tau$ ranking similarity metric.}
\label{fig:time-correlation}
\end{figure}

    In \autoref{fig:time-correlation}, we present the similarity measure between the generated rankings and the speeches' dates of delivery. We vary the hyperparameters, the model under analysis, and the dataset used. This enables a comparison of which configurations produce the most accurate temporal modeling. On the left, we can see that both models achieve significantly high results, with $\tau$ values above 0.8 for \texttt{voyage-lite-01} (note that a possible baseline could be 0, corresponding to the use of a random order). Additionally, the scores from \texttt{voyage-lite-01} achieve significantly and consistently higher correlation than those from \texttt{ada-002}, suggesting that \texttt{voyage-lite-01} models the temporal dimension more accurately. We also notice that, as we increase the $K$ value, the correlation between the scores and the date of the speeches increases consistently across both models. This indicates that the value of the $K$ hyperparameter affects the quality of the generated ranking, and that the comparison between models may be extrapolated for similar K values.
    
    On the right chart we observe that, when using only the speeches corresponding to a single party, the results change notably. Specifically, we see that \texttt{voyage-lite-01} performs better for the \textit{Conservative Speeches} (0.87), followed by the \textit{Presidential Speeches} (0.79), and finally the \textit{Democratic Speeches} (0.71). This would indicate that the \texttt{voyage-lite-01} model achieves similar results in terms of its ability to model the temporal dimension whether using all speeches or just those from a single party. Now, let's analyze the results obtained when using the \texttt{ada-002} model. We can see that it achieves a higher correlation for the \textit{Democratic Speeches} (0.80), followed by the \textit{Conservative Speeches} (0.72), and finally the \textit{Presidential Speeches} (0.59). This would suggest that, in the case of \texttt{ada-002}, better results are obtained when using speeches from a single party.
    Considering all of the above, we can conclude that \texttt{voyage-lite-01} has a greater ability to model the temporal dimension than \texttt{ada-002} for the case of Presidential and Conservative Speeches, but a lower capacity for Democratic Speeches. Overall, both models capture enough temporal information to achieve a significant correlation with the speech delivery date, confirming our hypothesis.

%IDEA DE ESTEBAN COMO QUEDARIAN LAS COSAS JUNTAS 

% ----------------------------------------------
% ---------------  Space  ----------------------
% ----------------------------------------------

\begin{figure}[htb]
 \begin{minipage}{.4\textwidth}
    \centering
    \includegraphics[width=\textwidth]{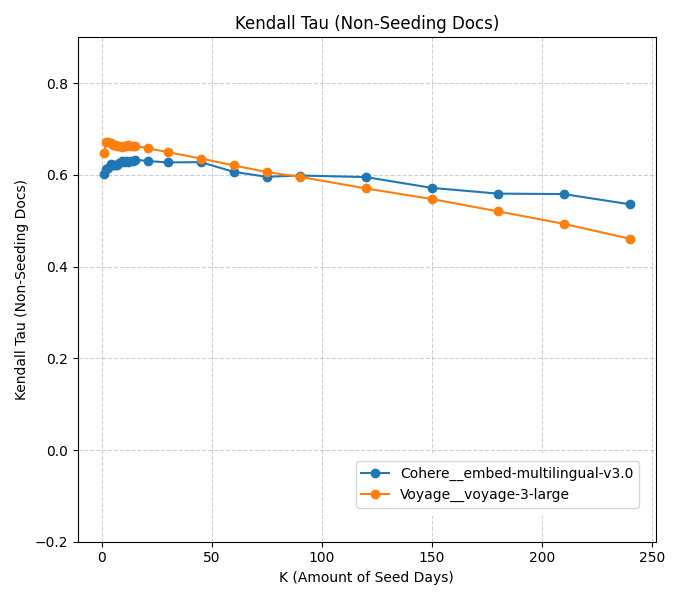}
    \Description{}
    \caption{\textit{Kendall $\tau$ vs. Amount of days used as seeds for \texttt{Extremes}.}}
    \label{fig:time-extremes}
\end{minipage}\hfill
 \begin{minipage}{.4\textwidth}
    \centering
    \includegraphics[width=\textwidth]{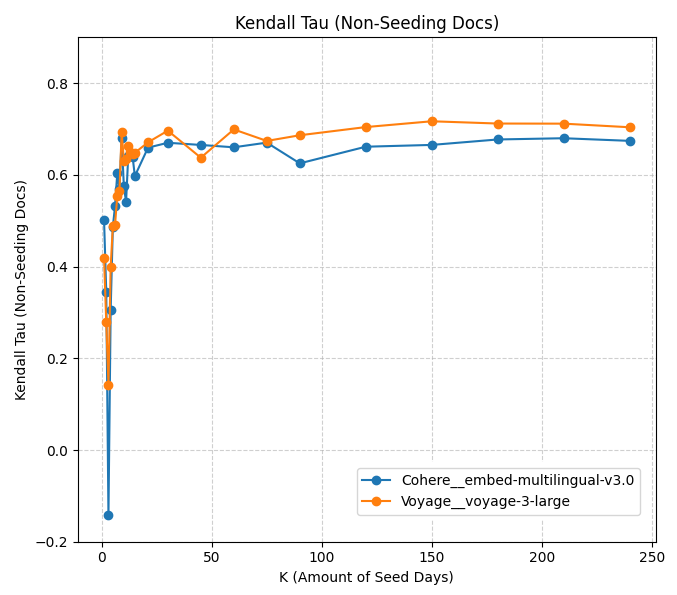}
    \Description{}
    \caption{\textit{Kendall $\tau$ vs. amount of days used as seeds for \texttt{Random Sorted}.}}
    \label{fig:time-randomsorted}
\end{minipage}
 \end{figure}

To deepen the research and test the limits of the temporal signal, we extended our analysis to a high-density dataset. 
Despite the increased granularity, we found that the temporal signal remains robust when using the Extremes Method for seed selection. All models achieved high correlations when projecting daily embeddings onto the temporal axis as seen in \autoref{fig:time-extremes} (where the projection of seed elements is not accounted for in the ranking metrics). 
The correlation was even better (for larger values of $K$) for the Random Sorted method, as seen in \autoref{fig:time-randomsorted} (we plot just two of the models for visual clarity).
By contrasting sampling strategies, we confirmed that the arrow of time is an intrinsic property distributed across the dataset: the Random Sorted method yielded significantly higher correlations than the Random Control.
We further discovered that interval length is critical. Through our chunking analysis, we observed that chunks containing more days systematically produce better ranking metrics than smaller chunks. As the data is better spaced in time, the temporal signal becomes clearer, enabling to smooth out daily noise (news topics) and recover the underlying linear progression of time.

\begin{figure}[!htb]
    \centering
    \includegraphics[width=0.525\textwidth]{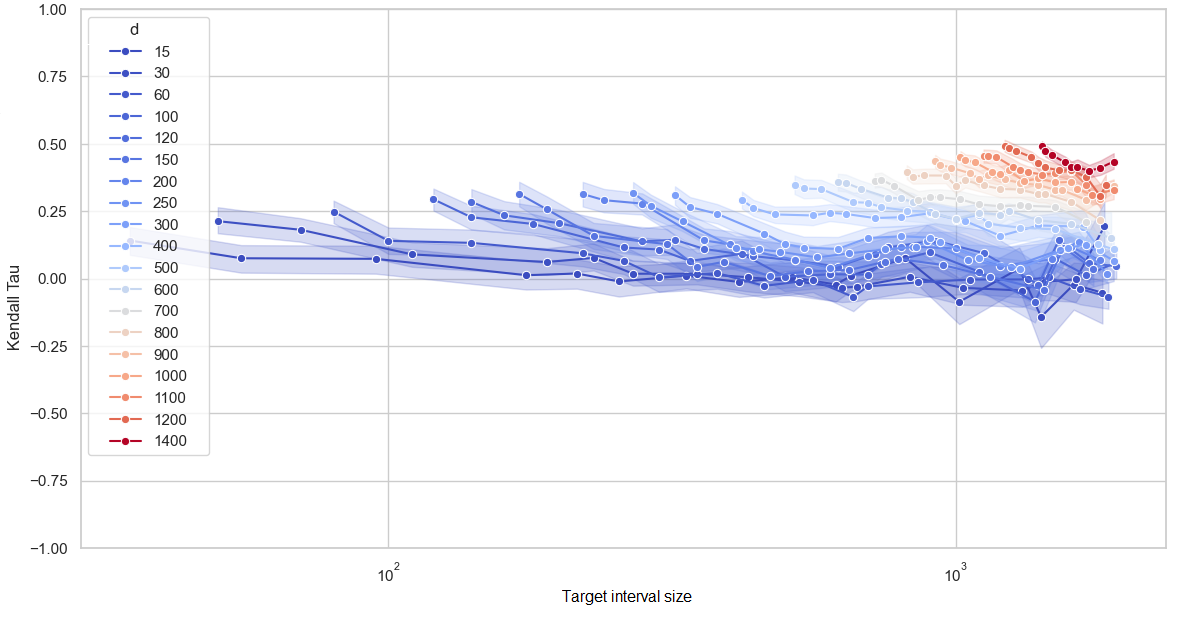}
    \Description{}
    \caption{For \textit{voyage-3-large, $K=3$ and 100 samples. Kendall $\tau$ vs. Target interval size to project. How close (in days) the extremes are from the target interval.}}
\label{fig:time-targetinterval}
\end{figure}

Another key finding is that defining the temporal axis locally (i.e. using seeds that are close to the target interval) yields better precision than using a global axis for the entire period. Performance degrades as the projection context expands too far beyond the target dates as seen in \autoref{fig:time-targetinterval}.
While the models excel at ordering dates between seeds (interpolation), they fail to reliably extrapolate: projections of future dates based on past axes do not follow a linear extension and the same applies for past days projections based on future axes. This suggests the learned time is a local relative ordering rather than an absolute magnitude, reflecting time-correlated content differences (semantic and thematic drift) rather than a universal, abstract “time” axis. The extracted temporal dimension is strictly local and tied to the progression of the events in that period.

% \begin{figure}[!htb]
%     \centering
%     \includegraphics[width=0.45\textwidth]{img/time/timeness-3-all.eps}
% \caption{\textbf{Varying models}. \textit{This chart shows how the timeness scores correlates with the actual time of delivery of the speeches for the models under analysis using K = 3.}}
% \label{fig:time-timeness-models}
% \end{figure}

% \begin{figure}[!htb]
%     \centering
%     \includegraphics[width=0.45\textwidth]{img/time/timeness-vlite-all.eps}
% \caption{\textbf{Varying K}. \textit{This chart shows how the timeness scores correlates with the actual time of delivery of the speeches using different values of the K hyperparameter for the \texttt{voyage-lite-01} model.}}
% \label{fig:time-timeness-k}
% \end{figure}

\subsection{Space Dimension}
\label{sec:space-exp}

    For the space dimension, we report results on two datasets. Although both focus on the goal of verifying how representation models reflect spatial coordinates, with the second one we focus on determining whether the model recognizes the circular nature of longitude: while latitude is reasonably modeled as a linear magnitude, longitude forms a continuous loop around the Earth, and we try to evaluate whether the model inherently captures this cyclical property or not.

\subsubsection{Datasets}

    \paragraph{WorldPlaces.} This dataset was curated to encompass a diverse selection of geographically and culturally significant locations, ensuring broad global representation. The names of the places were extracted from their respective Wikipedia pages, and they were primarily selected based on two criteria: their average Wikipedia page views over the past 9 years\footnote{\url{https://pageviews.wmcloud.org/}}
    and their status as UNESCO World Heritage sites\footnote{\url{https://whc.unesco.org/en/list/}}.
    Additionally, some locations were added to act as “extreme points” for our method (Svalbard Global Seed Vault and Golden Gate Bridge).

    \paragraph{PopulatedPlaces.} This dataset builds upon the \texttt{world\_place} dataset introduced by \cite{gurnee2023language}\footnote{The original dataset can be accessed at \url{https://github.com/wesg52/world-models/tree/main/data/entity\_datasets}}. While the original dataset includes a wide range of locations such as monuments, buildings, and restaurants, we apply a filtering criterion to include only populated places with more than 2.5 million \textit{page\_views}.

    It is important to highlight that latitude values exhibit greater clustering compared to longitude values. This difference arises from the fact that latitude is constrained within a range of -90\textdegree\ to 90\textdegree\, whereas longitude spans a broader range, from -180\textdegree\ to 180\textdegree. 
    %Note that we decided to use decimal degrees, which are decimal fractions used to represent latitude and longitude geographic coordinates.

\subsubsection{Results} 

\begin{figure}[htb]
\begin{minipage}{.5\textwidth}
    \centering
    \includegraphics[width=\textwidth]{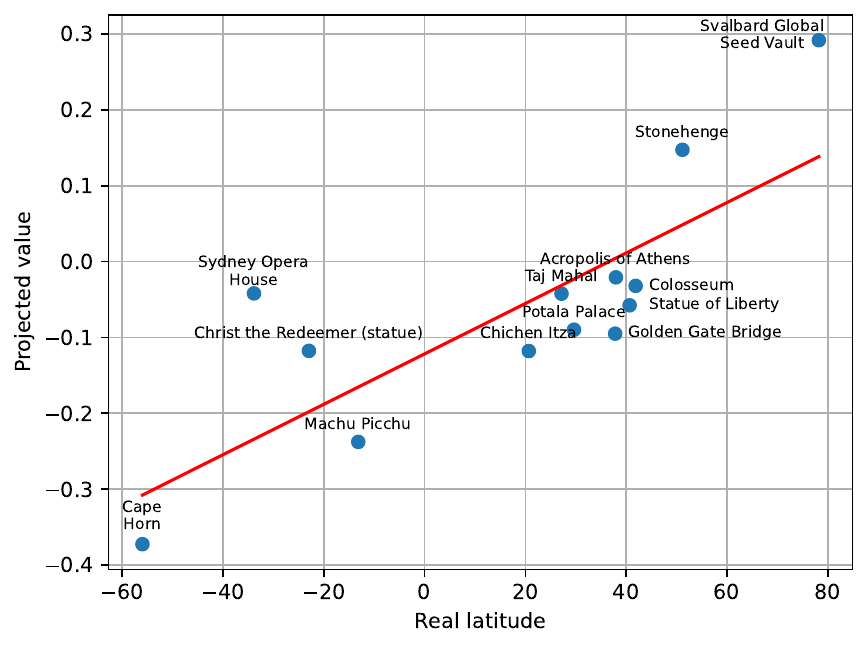}
\end{minipage}\hfill
\begin{minipage}{.5\textwidth}
    \centering
    \includegraphics[width=\textwidth]{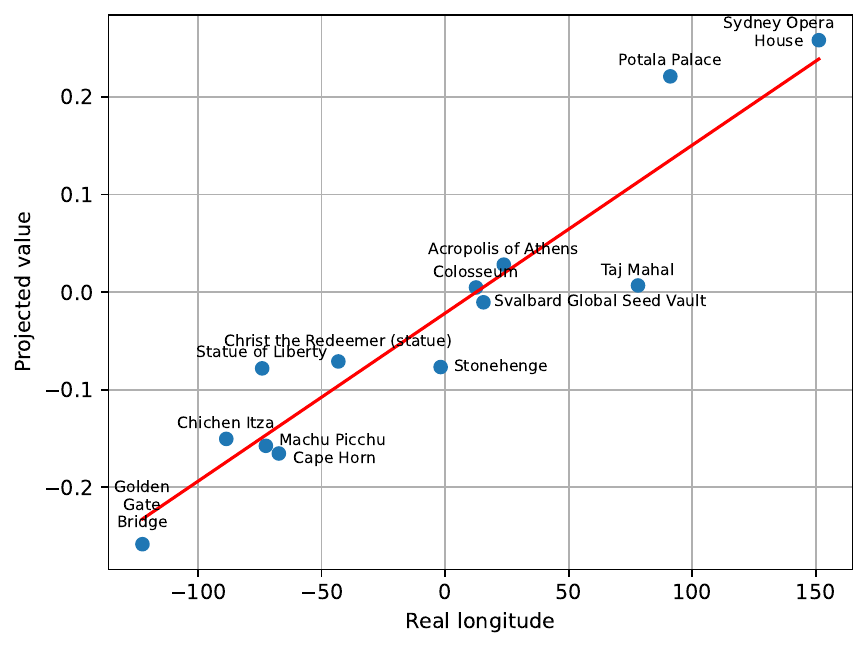}
\end{minipage}
\Description{}
\caption{Obtained values vs their real values. The upper graph shows the latitude while the lower graph shows the longitude. In both cases the model used was \texttt{text-embedding-004} and two pairs of seeds were used ($K = 2$).}
\label{fig:monumentsGemini004LatAndLong}
\end{figure}

    To generate the embeddings for the dataset \textit{WorldPlaces}, we simply use the names of the locations. In \autoref{fig:monumentsGemini004LatAndLong} we present an analysis of the score achieved by the algorithm versus the real value of the actual place. In the upper chart, we can see that there is a positive correlation between the real latitude and the projected value. The correlation values obtained were $\tau = 0.615$ and $r = 0.796$.  In the lower chart, we can also see a positive correlation, in this case between the real longitude and the projected value. The correlation values obtained were $\tau = 0.795$ and $r = 0.941$. 

Although our method does not yield precise values, we sought to examine whether augmenting the obtained latitude and longitude values and plotting them on a world map could provide an approximation to the actual geographic locations. For this, we employed scikit-learn's 
\href{https://scikit-learn.org/1.5/modules/generated/sklearn.preprocessing.MinMaxScaler.html}{\textit{MinMaxScaler}}, 
to scale each feature within the bounds of the minimum and maximum latitude/longitude values. This exploratory approach aimed to assess the potential for our method to approximate real-world coordinates, despite the inherent limitations in accuracy. 

\begin{figure*}[htb] 
\centering
\includegraphics[width=\textwidth]{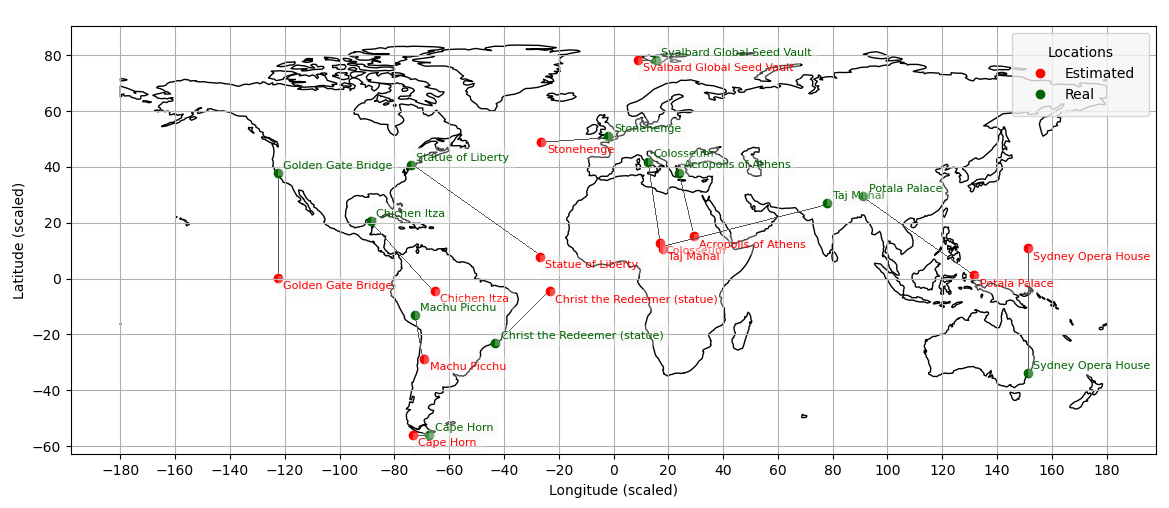}
\Description{}
\caption{Map constructed by scaling the values obtained on \autoref{fig:monumentsGemini004LatAndLong}.}
\label{fig:monumentsMap}
\end{figure*}

% Comentario del Mapa
As observed in \autoref{fig:monumentsMap}, while extreme values are determined by the scaling process, it is noteworthy that locations such as \textit{Cape Horn} or \textit{Svalbard Global Seed Bank}-despite being latitude extremes-are positioned nearly accurately on their corresponding longitudinal coordinates. 
Similarly, sites like \textit{Stonehenge} and \textit{Machu Picchu} are closely positioned to their actual locations. 
However, landmarks such as the \textit{Statue of Liberty} are significantly displaced. 
Two hypotheses for this discrepancy are the existence of numerous replicas of the Statue of Liberty worldwide\footnote{\url{https://en.wikipedia.org/wiki/Replicas\_of\_the\_Statue\_of\_Liberty}}, and 
the presence of common words such as \textit{statue} and \textit{liberty}, potentially influencing the model toward a more general location.
% Comentario del 2do Dataset ---------------------
% Tau:  0.8181818181818181
% Pearson:  0.9132018041925136    
% p-value: 3.349025959699737e-05
% Semillas utilizadas [('Tokyo', 'Manhattan'), ('Melbourne', 'Brooklyn')]

As we observe, the embeddings contain an underlying awareness of the geospatial locations. However, unlike latitude, longitude forms a continuous loop around the Earth. This brings us to a key question: Do models recognize this phenomenon? For the \textit{PopulatedPlaces} dataset, we focused exclusively on the longitude. In this instance, the lowest longitude was not used as the starting point. Instead, we chose to divide the dataset into three sections: from Manhattan to Budapest, from Budapest to Tokyo, and from Tokyo to Manhattan. The most notable aspect arises in the third section, which contains values spanning from positive to negative. 

\begin{figure}[!htb]
\centering
\minipage{0.45\textwidth}
    \centering \scriptsize
    \includegraphics[width=\linewidth]{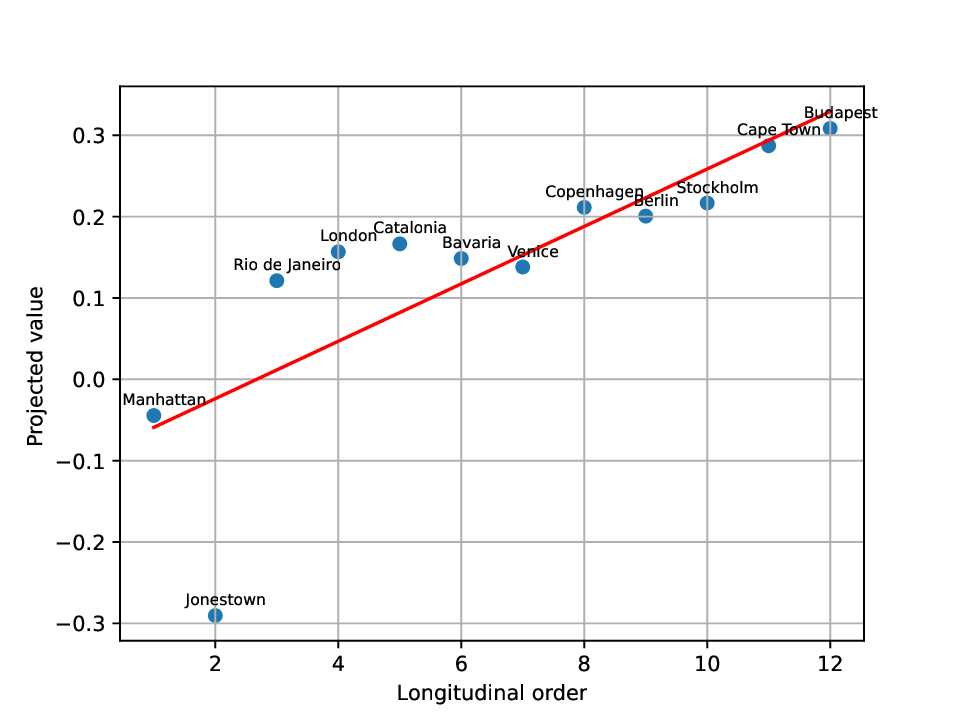}
    \textit{Manhattan–Budapest}
\endminipage\hfill
\minipage{0.45\textwidth}
    \centering \scriptsize
    \includegraphics[width=\linewidth]{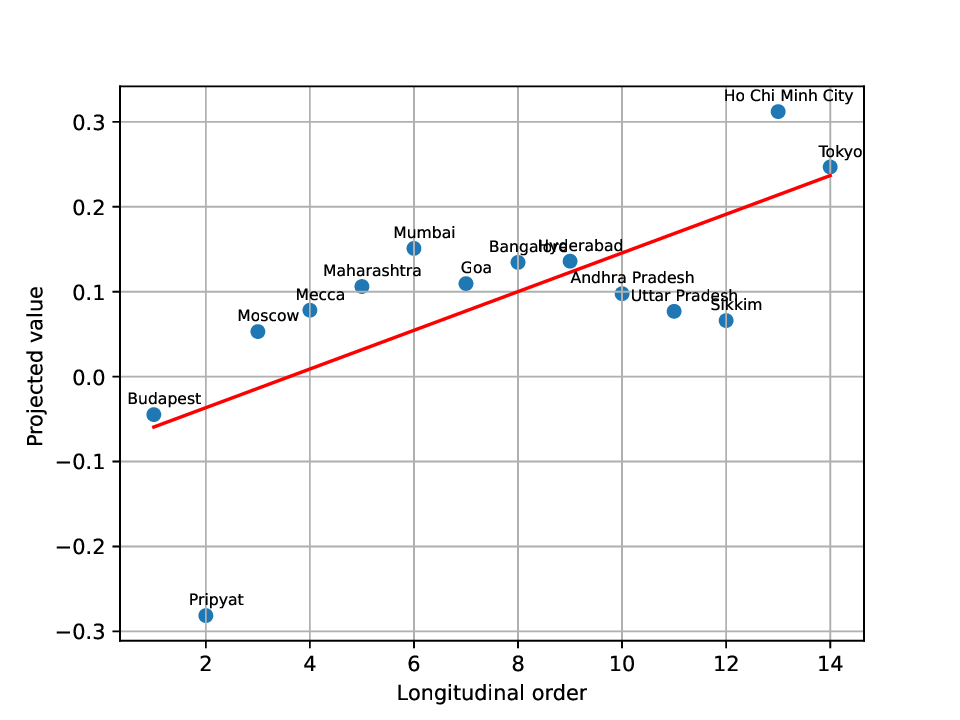}
    \textit{Budapest-Tokyo}
\endminipage\hfill
\minipage{0.45\textwidth}%
    \centering \scriptsize
    \includegraphics[width=\linewidth]{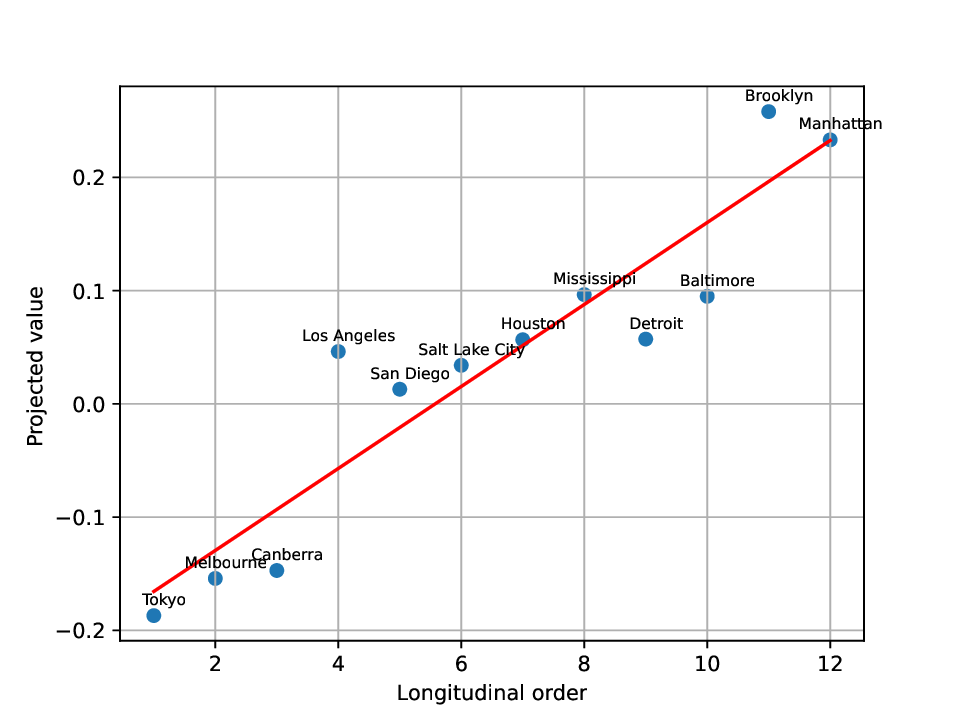}
    \textit{Tokyo–Manhattan}
\endminipage
\Description{}
\caption{Obtained values over their longitudinal order, for model \texttt{text-embedding-004}.}
\label{fig:citiesGemini004Long}
\end{figure}

As shown in \autoref{fig:citiesGemini004Long}, the two upper charts, while not perfect, exhibit the previously mentioned spatial order. The first chart demonstrates a reasonable representation ($\tau = 0.788$ and $r = 0.79$), whereas the second one shows a weaker correlation ($\tau = 0.451$ and $r = 0.7$).  We believe that this discrepancy may result from the proximity of the cities. 
In contrast, the lowermost chart presents a stronger correlation ($\tau = 0.848$ and $r = 0.931$), suggesting that the model may be capturing a circular representation of longitude. We are currently investigating an approach to improve longitude determination, as detailed in the following section. 

It is important to highlight that the  embeddings were constructed
using the name of the city and country  
(for example: ``Manhattan, United States"). 
Some cities were very close to one another, and for this reason, a simple number order on the x-axis was preferred to enhance readability, prevent excessive clustering of data points, and avoid confusion having positive values to the left of negative ones. 
We used a simple numerical order to calculate the values for Pearson's correlation. However, it is important to note that using the actual distances between the locations for the calculation results in nearly identical values.

\section{Discussion and Further Work}
\label{sec:conclusions}

We introduce a simple yet effective technique that operates solely on the embeddings produced by a chosen LLM. Our results provide further evidence that these embedding spaces encode grounded information about the real world, including temporal and geospatial structure.
More generally, the same projection-based procedure can be used to probe how embeddings capture other semantic concepts by selecting appropriate seed sets, offering a lightweight tool for interpretability. Finally, this capability may benefit information retrieval settings, for example by enabling unsupervised temporal or geographic tagging and by supporting dimension-aware ranking.

We focused in particular on the temporal and geographic dimensions. We found that attention-based language models, such as \texttt{voyage-lite-01} and \texttt{ada-002}, are capable of capturing information about the temporality of the textual units they process, encoded directly in the generated embeddings. This allows for the design of unsupervised methods that use only the embeddings to analyze the ability of the language model under study to capture the temporal dimension of the textual units. For example, the voyage-lite-01 model achieved Kendall's $\tau$ score of 0.87 on the Conservative Speeches dataset, indicating a high capacity to internally model the temporality of the textual units. 
This temporal dimension that can be extracted from the embeddings has a highly significant correlation with the release date of the speeches, reinforcing the validity of our approach. It is worth noting that while the creation time and focus coincide in our time datasets, further investigation is needed to explore the implications when they do not coincide.

The temporal granularity of these speeches and some other datasets we explored is yearly. We also considered datasets with daily granularity. 
Crucially, our analysis allowed us to test the robustness and limits of the time signal. We confirmed that the arrow of time is an intrinsic property distributed across the dataset, not merely an artifact of the extremes. By employing a \textit{Random Sorted} method (where seeds are selected randomly but assigned as early or late) we showed that the signal persists even without using the absolute extremes. Furthermore, regarding granularity, we observed a strong relationship between context size and signal clarity. Our partitioning analysis demonstrated that larger temporal windows allow the model to smooth out short-term semantic noise, resulting in significantly better linear ordering compared to smaller, fragmented windows.

We also identified that the representation is local rather than universal: while the method operates effectively as an interpolator, it struggles to extrapolate. 

Our results regarding the spatial dimension suggest that 
the embeddings encode information correlated with geospatial locations, and it appears that longitude is being represented in a non-linear manner. We are currently conducting further experiments to validate whether a circular embedding may emerge more clearly in that case through a circular, rather than linear, projection.

As for the explainability of our results, i.e., the questions ``how'' or ``why'' the temporal signal emerges, it is important to note that we are not claiming a unique reason for it. The recoverable temporal signal may depend on the semantic domain and the amount of supervision available, as well as on other factors. Topical drift, jargon, and grammatical variation are possible alternative explanations for the temporal signal. A similar issue appears in geography: cases such as ``Statue of Liberty'' suggest that embeddings may encode geographic coordinates from implicit or explicit sources. To further investigate this issue, we performed two kinds of experiments. First, we compared results with and without broader context: for example, dating book titles with and without the author's name, and locating places using embeddings of full Wikipedia articles instead of place names alone. These experiments suggest that broader context improves accuracy. Second, we ran additional controls masking explicit temporal references, such as years, events, and person names. The temporal signal still emerges under these controls, although these results remain preliminary.

Although we acknowledge that the proposed method requires further validation through additional experiments, the results obtained so far and reported here strongly suggest its effectiveness in addressing the target problem. 
Moreover, our approach was evaluated on additional datasets not included in this article for space reasons. The results obtained are conclusive: in all cases,  high correlations are observed between the measured dimension and the true underlying baseline values.
At this point, it is important to emphasize again that our goal was to test whether signals reported in prior probe or white-box-based work~\cite{DBLP:conf/iclr/EngelsMLGT25, gurnee2023language} can also be detected directly in final-layer embeddings. For that reason, we did compared our results directly against "ground truth" and not against methods such as TempoBERT which, as we said rely on explicit temporal supervision and are thus not directly comparable to our black-box setting. 

Additional experiments are been conducted on other domains and datasets. 
In particular, the projection method has been applied to a state-of-the-art symbolic music LLM, finding that it reveals signals relevant to music. We omit the reference for anonymity reasons. In that semantic domain we are currently testing certain circular attributes for the ideas proposed above for longitude. Preliminary results in this area are promising. 

Finally, we aim to explore fine-tuning mechanisms to better adjust temporal results, particularly addressing the different time dimensions (document creation time and focus time). 
Another research direction is related to some prompt engineering aspects~\cite{sahoo2024systematic}, namely the impact of different prompt formulations on the emergence of the  signals.

Overall, our findings suggest that temporal and geographic signals in language-model embeddings can be exploited not only for interpretability, but also as lightweight components for ranking, tagging, and metadata enrichment in retrieval-oriented systems.

\clearpage

\bibliographystyle{splncs04}
%el estilo de biblio debería ser ACM-Reference-Format, pero da error en cada referencia
%\bibliographystyle{ACM-Reference-Format}
\bibliography{thebibliography}

\end{document}